\documentclass[letterpaper]{article} %
\usepackage{aaai24}  %
\usepackage{times}  %
\usepackage{helvet}  %
\usepackage{courier}  %
\usepackage[hyphens]{url}  %
\usepackage{graphicx} %
\usepackage{natbib}  %
\usepackage{caption} %
\usepackage{algorithm}
\usepackage{algorithmic}
\usepackage{booktabs}
\usepackage{tabularx}
\usepackage{multirow}
\usepackage{graphicx}
\usepackage{adjustbox}
\usepackage{svg}
\usepackage[page,toc]{appendix}
\usepackage{makecell}

\usepackage{newfloat}
\usepackage{listings}
\DeclareCaptionStyle{ruled}{labelfont=normalfont,labelsep=colon,strut=off} %
\floatstyle{ruled}
\newfloat{listing}{tb}{lst}{}
\floatname{listing}{Listing}
\title{\textit{Scaling Crowdsourced Election Monitoring}:\\Construction and Evaluation of Classification Models for Multilingual and Cross-Domain Classification Settings}
\author {
    Jabez Magomere\textsuperscript{\rm 1},
    Scott A. Hale\textsuperscript{\rm 1, \rm 2}
}
\affiliations {
    \textsuperscript{\rm 1}University of Oxford\\
    \textsuperscript{\rm 2}Meedan\\
    \texttt{jabez.magomere@keble.ox.ac.uk}
}

\begin{document}

\maketitle

\begin{abstract}
The adoption of crowdsourced election monitoring as a complementary alternative to traditional election monitoring is on the rise. Yet, its reliance on digital response volunteers to manually process incoming election reports poses a significant scaling bottleneck. In this paper, we address the challenge of scaling crowdsourced election monitoring by advancing the task of automated classification of crowdsourced election reports to multilingual and cross-domain classification settings. We propose a two-step classification approach of first identifying informative reports and then categorising them into distinct information types. We conduct classification experiments using multilingual transformer models such as XLM-RoBERTa and multilingual embeddings such as SBERT, augmented with linguistically motivated features. Our approach achieves F1-Scores of 77\% for informativeness detection and 75\% for information type classification. We conduct cross-domain experiments, applying models trained in a source electoral domain to a new target electoral domain in zero-shot and few-shot classification settings. Our results show promising potential for model transfer across electoral domains, with F1-Scores of 59\% in zero-shot and 63\% in few-shot settings. However, our analysis also reveals a performance bias in detecting informative English reports over Swahili, likely due to imbalances in the training data, indicating a need for caution when deploying classification models in real-world election scenarios.
\end{abstract}

\section{Introduction}
The widespread availability of mobile telephony and the proliferation of social media platforms has given rise to ``liberation technologies'', enabling citizens to actively engage in democratic processes \cite{diamond_liberation_2010}.   These technological affordances have facilitated the emergence of crowdsourced election monitoring as a complementary alternative to traditional election monitoring \cite{okolloh_ushahidi_2009}. While the coverage and diversity of traditional election monitoring is constrained by the availability of trained electoral observers \cite{gromping_many_2012}, crowdsourced election monitoring addresses this limitation by harnessing the ``wisdom of the crowds'' \cite{surowiecki_wisdom_2005}--- i.e., a potentially infinite number of untrained observers who submit less structured anecdotal electoral observations \cite{gromping_many_2012}. 

\begin{figure}
    \centering
    \includegraphics[width=0.7\columnwidth]{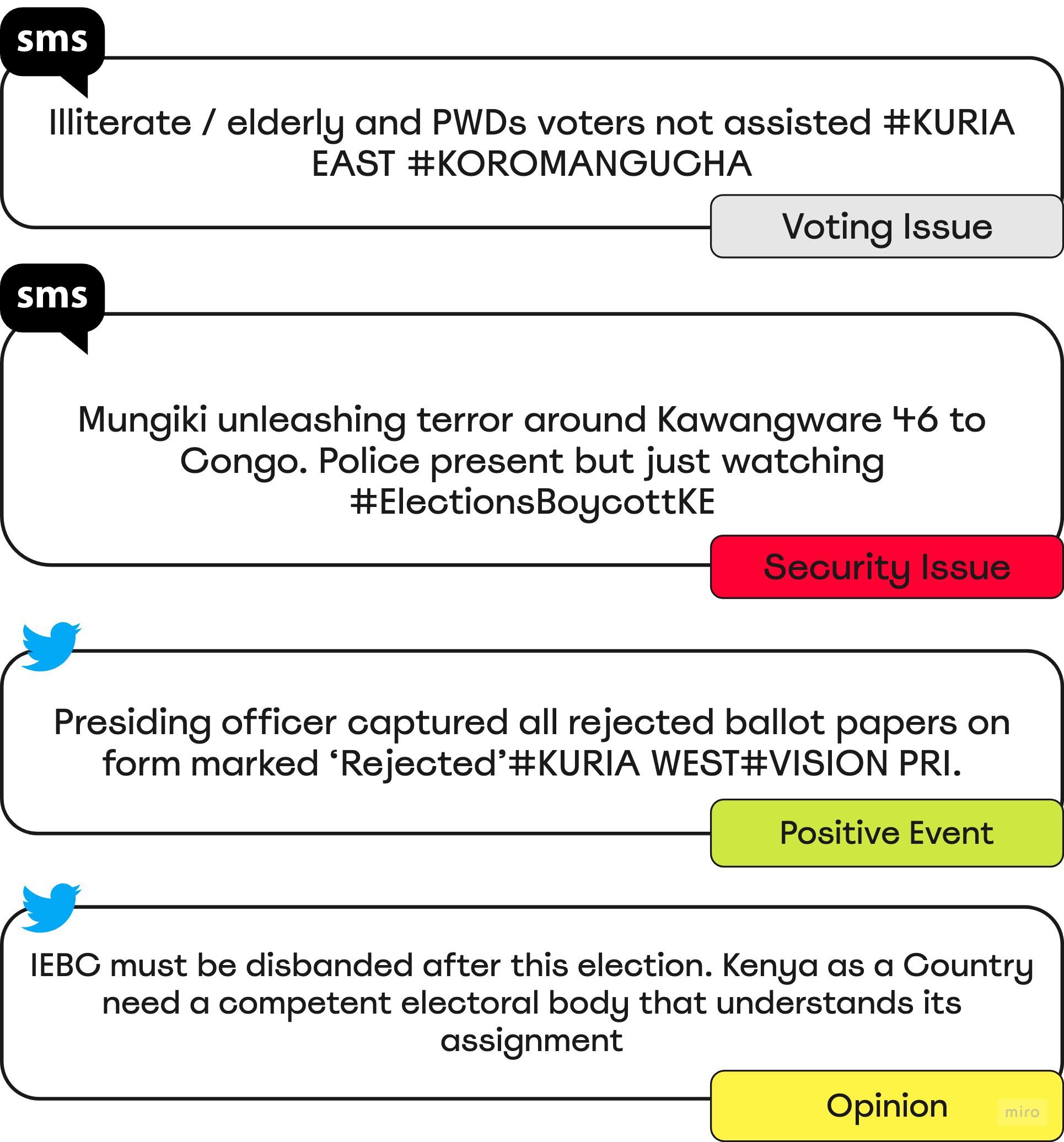}
    \caption{Examples of crowdsourced election reports and their annotated information types.}
    \label{fig:reports_example}
\end{figure}

The first application of crowdsourced election monitoring technologies was in the Kenyan 2007/2008 post-election violence, where, \textit{Ushahidi}\footnote{\url{https://www.ushahidi.com/}} (which means ``witness'' in Kiswahili) was created as a tool to allow citizens who witness acts of violence to report incidents via mobile phone messaging (SMS) or web forms \cite{okolloh_ushahidi_2009}. As of 2019,  over 19 crowdsourced election monitoring missions had been deployed in over 14 countries to monitor general elections with \textit{Ushahidi} being the most used software followed by \textit{Aggie}\footnote{\url{https://github.com/TID-Lab/aggie}} \cite{smyth_lessons_2016} and \textit{ELMO}\footnote{\url{https://electionstandards.cartercenter.org/1990/12/06/elmo/}} \cite{sassetti_social_2019}. Figure \ref{fig:reports_example} displays examples of crowdsourced reports from the 2022 Kenyan General Election.

Numerous studies have explored the positive impacts of crowdsourced election monitoring deployments, including detecting fraud and election malpractices \cite{bader_crowdsourcing_2013}, increasing voter turnout \cite{bailard_crowdsourcing_2014}, mitigating election-related violence \cite{trujillo_role_2014}, increasing political participation \cite{hellstrom_crowdsourcing_2015,shayo_crowdmonitoring_2017}, and improving election transparency \cite{sassetti_social_2019}. While the beneficial outcomes of crowdsourced election monitoring are widely acknowledged in scholarly discourse, the reliance on human annotators for processing incoming reports remains a significant scaling bottleneck \cite{smyth_lessons_2016, shayo_crowdsourcing_2017}. 
Similar to the challenges observed in traditional election monitoring \cite{gromping_many_2012}, the scarcity of human resources available to process the influx of crowdsourced election reports results in a high number of events going unreported.

As a response to these challenges, \citet{meier_automatically_2013} proposed the task of automated classification of crowdsourced election reports in the Artificial Intelligence for Monitoring Elections (AIME) project\footnote{\url{https://irevolutions.org/2013/04/17/ai-for-election-monitoring/}}. However, the use of these classification models in crowdsourced election deployments remains limited. Conventional approaches to this task train classification models on a limited dataset of English-only reports, develop election-specific models and evaluate the models solely based on accuracy without accounting for performance across language groups. Models constructed within such frameworks exhibit limited practical relevance to real-world election deployments, highlighting a gap between the development and deployment of such models. In this work, we aim to bridge this gap by advancing the task of automated classification crowdsourced election reports to multilingual settings and cross-domain election deployment scenarios while accounting for model performance across language groups.

In this study, we propose a two-step classification approach of first identifying informative election reports and then classifying them into information types relevant to the different electoral stakeholders. We achieve promising results on both classification tasks using state-of-the-art multilingual models augmented with linguistically motivated features. In our experiments, we demonstrate that models trained on crowdsourced electoral reports from one electoral domain can be effectively transferred to new electoral domains with minimal or no additional training. While these results are encouraging, we observe performance disparities in detecting informative election reports in English compared to Swahili. This highlights the need for a more balanced language group representation in the training data for effective real-world application by election monitoring organisations. We also conduct a thorough error analysis and suggest that future crowdsourced election monitoring organisations adopt better annotation strategies to prevent ambiguous information types that negatively impact model performance.

\section{Related Work}

Classifying crowdsourced election reports is typically modelled as a binary task of detecting election-related reports \cite{yang_using_2016, sambuli_viability_2013, muchlinski_we_2021} or a multi-class classification task that classifies election reports into information categories \cite{meier_automatically_2013, meier_17_2015}. \citet{meier_automatically_2013} conducted the first study on automating the classification of crowdsourced election reports utilising crowdsourced reports from the Kenya 2013 General Election sourced from \textit{Ushahidi}. The study explored gradient-boosting techniques on unigrams and bigrams extracted from the election report's text and title represented using term frequency-inverse document frequency (TF-IDF) scores. The models exhibited low recall in identifying sensitive information categories such as violence reports (24.8\%) and voter issues (57.3\%).  In subsequent work, \citet{meier_17_2015} extended this approach to the Nigerian 2015 General Election where classification models are trained on 571 human-tagged tweets to detect election violence, rigging, and voter issues.  The classifiers achieve low precision in identifying voting issues (47\%) and low recall in detecting violence reports (33\%).

While not directly related to classifying crowdsourced election reports, \citet{sambuli_viability_2013}'s study offers relevant insights. They trained a Support Vector Machine (SVM) classifier on tweets represented as TF-IDF vectors from the 2013 Kenyan Election to determine newsworthiness, achieving 85\% precision. \citet{yang_using_2016} experimented with Word2Vec \cite{mikolov_efficient_2013} word embeddings and Convolutional Neural Networks (CNNs) on the task of identifying election-related tweets, using 5,747 Spanish tweets from Venezuela's 2015 Parliamentary Election. Their approach, using word embeddings with CNNs, achieved an F1-Score of 77\%, surpassing TF-IDF-based SVMs. In a related study, \citet{muchlinski_we_2021} also utilised CNNs with word embeddings for classifying tweets containing electoral violence. They trained binary classifiers for each electoral context using tweets from the Venezuela 2015 Election (5,747 tweets), the Philippines 2016 Election (4,163 tweets), and Ghana 2016 Election (3,235 tweets), with F1-Scores ranging from 74\% to 77\%. These studies demonstrate potential in automating the classification of crowdsourced election reports. However, it is crucial to acknowledge and address the following outlined limitations:
\begin{itemize}
    \item \textbf{Limited Datasets:} Developing high-quality labelled datasets for election-related classification tasks can be time-consuming and expensive. To overcome this challenge, previous studies employed annotators to label a sample of the crowdsourced data to be used for model learning. However, this process introduces annotation biases \cite{olteanu_social_2019}, as the annotation guidelines used in post-election studies often differ from those employed in real-time election monitoring deployments. 
    Slight nuances in how annotation guidelines are written can also result in model biases and hurt model fairness \cite{hansen_guideline_2021}. 
    As such,  models developed using limited and imbalanced datasets typically exhibit low recall on sensitive information types like Voting Issues and Security Issues, as observed in \citet{meier_automatically_2013} and \citet{meier_17_2015}.
    
    \item \textbf{Cross Domain Classification Settings:} Previous approaches have largely focused on developing election-specific classifiers which limit their transferability to future elections, both within the same electoral domain (in-domain) or to new electoral domains (cross-domain). In the in-domain setting, classifiers are developed using crowdsourced election reports from a specific election, such as the Kenya 2022 General Election, for both training and testing. While this setup is relatively easier to study, it lacks practical relevance. On the other hand, the cross-domain setting, which is more realistic, involves training classifiers with crowdsourced election reports from one or more individual elections and testing them on reports from different elections. However, this setting requires more data and is, therefore, less studied.

    \item \textbf{Linguistic Disparities:} Previous research on classifying crowdsourced election reports has primarily focused on English datasets, except for \citet{yang_using_2016} and \citet{muchlinski_we_2021}, who developed classifiers for Spanish election tweets. Ironically, crowdsourced election monitoring technologies are mostly deployed in the global south \cite{gromping_many_2012, sassetti_social_2019} where online conversations frequently occur in non-English languages. Consequently, models trained on linguistically non-diverse datasets may exhibit performance disparities across language groups in real-world settings \cite{blasi_systematic_2021}.
\end{itemize}

In light of the aforementioned limitations, the adoption of classification models in crowdsourced election deployments has been constrained. In this work, we seek to overcome these constraints by advancing the task of classifying crowdsourced election reports to multilingual and cross-domain settings while also evaluating model performance across different language groups. 

\section{Dataset}
We obtain 239,301 crowdsourced election reports from \textit{Uchaguzi}\footnote{\url{https://uchaguzi.or.ke}} platform, a customised version of \textit{Ushahidi} deployed for monitoring the Kenyan General Elections in 2017 and 2022 and \textit{Uzabe},\footnote{\url{https://uzabe.ushahidi.io}} a customised deployment used in the 2023 Nigerian Election \footnote{Data can be requested from Ushahidi by contacting \\ \url{support@ushahidi.com}}. The deployments enable trained observers and the general public to actively participate in election monitoring by submitting observations through various communication channels, including WhatsApp, SMS (Short Message Service), USSD (Unstructured Supplementary Service Data), and Twitter. Additionally, the election deployments passively source election reports from social networking platforms such as Twitter by monitoring election-specific hashtags including \textit{\#Kenya\-Decides\-2022} and \textit{\#Nigeria\-Decides\-2023}. In all deployments, situation rooms were set up with teams of digital response volunteers responsible for processing incoming reports, assessing report credibility, translating, geo-locating, categorising, and mapping approved reports for public viewing on a crowd mapping platform \cite{shayo_crowdsourcing_2017}. The list of information categories and their descriptions used by \textit{Uchaguzi} is provided in Appendix \ref{subsec:info_categories}.

To construct a dataset suitable for the classification tasks, we use a combination of crowdsourced election reports from the 2017 Kenyan General Election and the 2022 Kenyan General Election deployments. We remove all unlabelled election reports, duplicated reports and reports with missing texts. We also exclude election reports sourced from USSD surveys as they provide users with a structured form to submit observations and as such the information types are applied by the end user and verified by a response volunteer. This results in a dataset of 14,181 labelled election reports. 

For the cross-domain classification experiments, we use crowdsourced election reports sourced from the 2023 Nigerian Election. Following similar steps applied to the Kenyan Election reports, we filter out all duplicated and unlabelled election reports, resulting in a dataset of 1,051 labelled election reports. 

\subsection{Classification Tasks}
To classify the reports, we adopt a two-step classification approach widely applied in crisis informatics  \cite{nguyen_applications_2016,ashktorab_tweedr_2014,alharbi_classifying_2022}. The first step involves filtering election reports based on their informativeness to remove noisy reports. The second step applies a fine-grained classifier to classify informative election reports into relevant information types.
 
\subsubsection{Informativeness Detection}
This task aims to identify informative election reports from crowdsourced data by classifying them as either informative or non-informative. This task is modelled as a binary classification task. We define informative election reports as: (1) reports that enhance understanding of the ground-level situation \cite{olteanu_what_2015,shaw_sharing_2013,sreenivasan_tweet_2011} and (2) reports that provide actionable and tactical information enabling decision-making, and guiding information-seeking processes \cite{vieweg_microblogging_2010} for the relevant electoral stakeholders. 
We identify the following categories as informative; \textit{Voting Issues}, \textit{Security Issues}, \textit{Polling Station Issues}, \textit{Staffing Issues}, \textit{Counting and Results Issues} and \textit{Positive Events}. Election reports categorised as \textit{Opinions} or \textit{Others} are considered non-informative.

\subsubsection{Information Type Classification} The objective of this task is to classify informative election reports into distinct information types that are relevant for various electoral stakeholders. This task is formulated as a multi-class classification problem. 
We combine \textit{Voting Issues}, \textit{Staffing Issues}, and \textit{Polling Station and Administration Issues} into a single category---\textit{Voting Issues}---, due to their conceptual similarities. This results in a 5-way classification task distinguishing between: \textit{Political Rallies}, \textit{Positive Events}, \textit{Security Issues}, \textit{Voting Issues}, and \textit{Counting and Results}. Election reports labelled as \textit{Media Reports} are excluded from both classification tasks as they are crowdsourced from a list of verified media entities and automatically assigned this label regardless of the report's content. 

The Nigerian election deployment uses a different set of information types than the other deployments. Information types with the same name are mapped one-to-one with those in the Kenyan dataset, such as \textit{Positive Events} and \textit{Security Issues}. The other information types are mapped based on their content similarity. For example, \textit{Sorting, Counting, and Collation} issues are mapped to \textit{Counting and Results Issues}, while \textit{Ballot Issues} and \textit{Polling Station Administration Issues} are mapped to \textit{Voting Issues}. The Nigerian deployment did not label any election reports as \textit{Political Rallies}, so the final set of information types for the classification task consisted of \textit{Positive Events}, \textit{Security Issues}, \textit{Counting and Results}, and \textit{Voting Issues}. A summary of the number of crowdsourced reports in each of the information types is shown in Table \ref{tab:summary_count}. 

\begin{table}[h]
\footnotesize
\centering
\setlength{\tabcolsep}{5pt} %
\renewcommand{\arraystretch}{1.2} %
\begin{tabular}{@{}lcc@{}}
\toprule
\multicolumn{1}{l}{\textbf{Classification Task}} & \textbf{Kenya} & \textbf{Nigeria} \\ \midrule
\textbf{Informativeness Detection}    &                &                  \\
\textit{Non-Informative}           & 9,304          & N/A              \\
\textit{Informative}               & 4,877          & 1,051            \\
\addlinespace
\textbf{Information Type Classification}                 &                &                  \\
Political Rallies                  & 387            & N/A              \\
Voting Issues                      & 1,141          & 608              \\
Counting \& Results                & 1,023          & 63               \\
Positive Events                    & 1,715          & 327              \\
Security Issues                    & 611            & 53               \\ 
\bottomrule
\end{tabular}
\caption{Summary of the number of crowdsourced reports per label in the classification dataset (Kenya) and the cross-domain dataset (Nigeria)}
\label{tab:summary_count}
\end{table}

The Kenyan election dataset contains 84\% (11,961 out of 14,181) of the crowdsourced reports in English, while 16\% (2,220 out of 14,181) are in non-English languages such as Kiswahili and Sheng'.\footnote{Sheng' is a hybrid language or slang that combines elements of English, Swahili, and local Kenyan dialects.} The Nigerian election dataset contains 97\% English reports and 3\% non-English reports. We posit that the small proportion of non-English reports observed is possibly attributed to the fact that a significant number of passively crowdsourced election reports that are more likely to contain non-English languages are not sufficiently labelled during the deployments. 

\section{Methods}
We train binary classifiers for the informative detection task and multi-class classifiers for the information type classification task. The data is split into stratified 70/10/20 train\slash dev\slash test splits for each classification task.

\subsection{Models}
\subsubsection{Baseline Models}
For the baseline experiments, we compare three widely used traditional machine learning models for text classification: Logistic Regression, Support Vector Machines, and Multinomial Naive Bayes. After preprocessing the election report text, we explore the use of unigrams, bigrams and a combination of both, represented as TF-IDF vectors. After initial experiments, we select Support Vector Machines as a reasonable baseline (see Appendix Section \ref{sec:tfidf_appendix} for results).
\subsubsection{Multilingual Sentence Transformers} We use the \texttt{paraphrase-multilingual-mpnet-base-v2}\footnote{At the time of conducting this study, this was the best-performing multilingual embedding model, see model leaderboard: \url{https://www.sbert.net/docs/pretrained_models.html}} model, a Sentence-BERT (SBERT) \cite{reimers_sentence-bert_2019} multilingual embedding model trained on over 50 languages including Swahili. Following a similar approach to \citet{reimers_sentence-bert_2019}, we use the election reports embeddings as inputs to a Logistic Regression classifier. We refer to this model as LR \textsubscript{SBERT}.
\subsubsection{Multilingual Transformer Models}
We evaluate the performance of XLM-R \cite{conneau_unsupervised_2020}, a multilingual masked language model trained on 2.5 TB of CommonCrawl\footnote{\url{https://data.statmt.org/cc-100/}} data in 100 languages. We also asses mDeBERTav3 \cite{he_debertav3_2021}, a multilingual version of DeBERTaV3, pre-trained on a similar dataset (CC100) as the XLM-R model. Based on the comparisons with other multilingual transformer-based models, XLM-R is selected for subsequent experiments as it exhibited slightly better performance than the other models (see Appendix \ref{sec:bert_appendix}).

\subsection{Additional Features}
We explore the effect of including the context, temporal information and sentiment polarity scores of each election report as additional features to the classification models. 
\subsubsection{Context} 
The information type an election report belongs to could be influenced by what is just happening before that report. For instance, if electoral identification kits fail at a particular time and place, the crowdsourced election reports from that period tend to include related information types. We capture the context of an election report by including the 3 previous election reports received before the current one as ordered by the timestamp in the dataset. To prevent the test set from leaking into the training and validation set, the context for each election report in the training set is only obtained from the training set while the test set has access to the full dataset.
\subsubsection {Temporal Information}
Crowdsourced election reports exhibit temporal dependencies with different information types received at different times. For instance, \textit{Voting Issues} are submitted a few days before the election and peak on the actual election day when people cast their votes as shown in Appendix Section \ref{sec:dist_appendix}. We include the absolute number of days between the official election date and the date each report was received as a feature in our analysis. To represent the time of day an election report was received, we apply cyclic transformations to the hour of the day such that each hour is represented by a cosine and sine function with values ranging from -1 to 1.
\subsubsection{Sentiment Features} Inspired by the work of \citet{verma_natural_2011} that showed that texts that are formal, objective and impersonal are more likely to contain situational awareness information, we include the sentiment scores---i.e., positive, negative and neutral scores---for each election report obtained using an XLM-R model\footnote{\url{https://huggingface.co/cardiffnlp/twitter-xlm-roberta-base-sentiment}}  \cite{barbieri_xlm-t_2021} fine-tuned for sentiment analysis. 

\noindent For the BERT-based models, we concatenate the context of an election with the election report text separated by a \texttt{[SEP]} token before being passed to the transformer layers, while the temporal and sentiment features are combined with the final BERT \texttt{[CLS]} embedding representation of the election report and its context. This combined vector is then passed to the final classifier. We experiment with different architectures for the final classifier, including a linear classifier and a dense multi-layer perception network. A linear classifier with dropout and batch normalisation achieves the best results. For the sentence embedding classifiers, we map each election report in the context to a 768-dimensional vector and average them to form one contextual embedding that is added as a feature input to the logistic classifier together with the election report's text embedding and the numerical features. 

\subsection{Cross-Domain Classification Settings}
We evaluate the performance of classification models trained on the Kenyan electoral domain when applied to the Nigerian electoral domain, representing a cross-domain classification setting. We focus on the information types classification task as the Nigerian dataset does not contain any non-informative election reports for the informativeness detection task. We conduct a zero-shot and few-shot experiment aimed at stimulating different crowdsourced election monitoring technologies deployments. 
\subsubsection{Zero-Shot Classification Setting}
In the zero-shot classification setting, classifiers trained on the Kenyan election dataset are directly applied to the Nigerian dataset without any exposure to examples from the Nigerian election domain hence zero-shot. This experiment simulates a scenario where a deployment is made in an electoral domain without any training instances. We compare the best-performing models from the Kenyan classification tasks: Logistic Regression with SBERT sentence embeddings (LR \textsubscript{SBERT}) and XLM-R.
\subsubsection{Few-Shot Classification Setting} We experiment with fine-tuning the models with a small number of reports sampled from the Nigerian election dataset to serve as training data (10\% or 105 election reports), while the remaining is used as a hold-out set, hence few-shot. This experiment simulates a scenario where crowdsourced election monitoring technologies are deployed to new contexts with minimal available training data at the onset of an electoral cycle. We compare three different models in this setting; (1) Logistic Regression with SBERT embeddings model (2) an XLM-R model finetuned on the Kenyan election dataset further finetuned with reports from the Nigerian election dataset and (3) an XLM-R model finetuned with reports only from the Nigerian election dataset.  

\section{Results}
We report results derived from averaging model performance across multiple runs \((n=3)\) with different randomly initialised train and test splits for each run. 

\subsection{Informativeness Detection}
Results in Table \ref{tab:informativeness_detection_results} show that all models perform relatively well on the task of detecting informative crowdsourced election reports. The XLM-R model fused with additional features achieves the highest F1 score at 77.54\%. However, it achieves this through improved precision at the expense of a lower recall,  a trade-off that may not be suitable for crowdsourced election monitoring where accurately identifying informative election reports is important.
Across all models, including the context, temporal information, and sentiment scores improve model performance.

\begin{table}[h]
\small
\centering
  \begin{tabular}{@{}llllll@{}}
    \toprule
    \textbf{Model}                   & \textbf{Acc. (\%)} & \textbf{P (\%)} & \textbf{R (\%)} & \textbf{F1 (\%)} \\
    \midrule
    SVM \textsubscript{TF-IDF}               & 79.39            & 73.69              & \textbf{79.04}          &   75.16 \\
    LR \textsubscript{SBERT}          & 79.61         & 75.47          & 78.05       & 76.42  \\
    LR \textsubscript{SBERT + features} & 80.21         & 76.29          & 78.70       & 77.20 \\
    XLM-R \textsubscript{text only}                 & 80.10             & 78.35         & 76.55       & 77.25 \\   
    XLM-R \textsubscript{text +  features}      & \textbf{80.25}         & \textbf{78.47}          & 76.90       & \textbf{77.54}     \\
    \bottomrule
  \end{tabular}
  \caption{Accuracy metrics for informativeness detection task (Maximum columnar values in bold)}
  \label{tab:informativeness_detection_results}
\end{table}

\subsection{Information Types Classification}
Results shown in Table \ref{tab:information_types_results} suggest that the task of classifying informative election reports into relevant information categories is more challenging compared to detecting informative election reports. The LR \textsubscript{SBERT + additional features} is the best-performing model with an F1-Score of 74.55\%, which is a 5-point improvement over the SVM baseline. Interestingly, LR \textsubscript{SBERT + additional features} (74.55\%) outperforms the fine-tuned XLM-R models (72.27\%) in terms of F1-Score. Similar to the informativeness detection task, including the context, temporal features and sentiment polarity scores resulted in performance improvements across all models with a notable 3 percentage-point improvement in terms of F1-Score for the LR \textsubscript{SBERT} model.

\begin{table}[hbt!]
  \small
  \centering
  \begin{tabular}{@{}llllll@{}}
    \toprule
    \textbf{Model}                   & \textbf{Acc. (\%)} & \textbf{P (\%)} & \textbf{R (\%)} & \textbf{F1 (\%)} \\
    \midrule
    SVM \textsubscript{TF-IDF}               &  70.05           & 70.01            & 
    69.31 &   69.54           \\
    LR \textsubscript{SBERT}          & 73.32        & 70.92          & 72.98     & 71.83           \\
    LR \textsubscript{SBERT + features} & \textbf{75.41}         & \textbf{73.71}         & \textbf{75.76}       &  \textbf{74.55}          \\
    XLM-R \textsubscript{text only}                 & 73.70             & 72.66        & 71.98       & 72.23        \\
    XLM-R \textsubscript{text + features}      & 72.91        & 71.58          & 73.83       & 72.27        \\
    \bottomrule
  \end{tabular}
  \caption{Accuracy Metrics for Information Types Classification (Maximum columnar values in bold)}
  \label{tab:information_types_results}
\end{table}

\begin{table*}[hbt!]
\centering
\adjustbox{max width=\textwidth}{
\begin{tabular}{@{}l*{15}{c}@{}}
\toprule
\multirow{3}{*}{}    & \multicolumn{6}{c}{\textbf{Zero-Shot Classification}}                                                                                                                         & \multicolumn{9}{c}{\textbf{Few-Shot Classification}}                                                                                                                                                                                        \\ \cmidrule(lr){2-7} \cmidrule(lr){8-16}
                     & \multicolumn{3}{c}{\multirow{2}{*}{KE-LR \textsubscript{SBERT}}} & \multicolumn{3}{c}{\multirow{2}{*}{KE-XLM-R}}       & \multicolumn{3}{c}{\multirow{2}{*}{NG-LR \textsubscript{SBERT}}}  & \multicolumn{3}{c}{\multirow{2}{*}{NG-XLM-R}} & \multicolumn{3}{c}{\multirow{2}{*}{KE-XLM-R}} \\ 
                     & \multicolumn{3}{c}{}                          & \multicolumn{3}{c}{}                               & \multicolumn{3}{c}{}                             & \multicolumn{3}{c}{}                  & \multicolumn{3}{c}{}                                     \\ \cmidrule(lr){2-4} \cmidrule(lr){5-7} \cmidrule(lr){8-10} \cmidrule(lr){11-13} \cmidrule(lr){14-16}
                     & F1  & P   & R   & F1           & P            & R            & F1           & P   & R   & F1  & P   & R   & F1           & P            & R            \\ \midrule
Counting and Results & .29 & .21 & .44 & .35          & .29          & .43          & .30          & .26 & .35 & .00 & .00 & .00& \textbf{.44} & \textbf{.35} & \textbf{.60} \\ 
Positive Events      & .65 & .71 & .60 & \textbf{.73} & \textbf{.70} & \textbf{.76} & .69          & .66 & .73 & .59 & .56 & .63 & .72          & .69          & .74          \\ 
Security Issues      & .41 & .36 & .47 & .51          & .54          & .49          & \textbf{.56} & .56 & .56 & .47 & .41 & .54 & .55          & \textbf{.60} & \textbf{.50} \\ 
Voting Issues        & .76 & .79 & .74 & .79          & .83 & .75          & .79          & .83 & .75 & .74 & .74 & .74 & \textbf{.80}          & \textbf{.85} & \textbf{.76} \\ \cmidrule{1-16}
Macro Avg            & .52 & .56 & .53 & .59          & .59          & .61          & .59          & .58 & .60 & .45 & .43 & .48 & \textbf{.63} & \textbf{.62} & \textbf{.65} \\ \bottomrule
\end{tabular}
}
    \caption{Models Performance in Zero-Shot and Few-Shot Cross-Domain Classification Settings (Maximum columnar values in bold)}
    \label{tab:cross-domain-results}
\end{table*}

\subsection{Cross Domain Classification Settings}
We examine the performance of models trained on one election domain dataset and transferred to a new target election domain in zero-shot and few-shot classification settings. Results of the cross-domain experiments are shown in Table~\ref{tab:cross-domain-results}.
\subsubsection{Zero-shot classification setting} Both models perform relatively well in the zero-shot setting given that the models have not been adapted to fit the new electoral domain. The XLM-R model (KE-XLM-R) achieves the highest performance with an F1-Score of 59\%.
\subsubsection{Few-shot classification setting}  The KE-XLM-R (F1-Score of 63\%), originally trained on the Kenyan election dataset and further fine-tuned for the Nigerian classification task, outperforms the domain-specific XLM-R model, NG-XLM-R (F1-Score of 45\%), trained entirely from scratch on the same amount of training instances. Surprisingly, the KE-XLM-R model applied in the zero-shot classification setting outperforms a domain-specific XLM-R model trained from scratch on the Nigerian dataset in the few-shot classification setting. The KE-XLM-R model, trained on a larger dataset, potentially acquired broader knowledge relevant to both Kenyan and Nigerian election reports. Notably, the NG-XLM-R model struggles to identify information types like \textit{Counting and Results}, likely due to their scarcity in the Nigerian training dataset.

\subsection{Performance Across Language Groups}
We compare the performance of the classification models between English  \((n=2,388)\) and Swahili \((n=87)\) election reports in the holdout set. 
Table \ref{tab:langauge_fairness_1} shows that the model performs better in identifying informative English election reports than Swahili ones, with a performance gap of 7 percentage points. 
We hypothesise that the differences in model performance between the two language groups can be attributed to imbalances in the training data, with the model being trained on 9,573 English reports compared to only 338 Swahili reports. We also note that the test set for Swahili is small.

\begin{table}[h]
  \small
  \centering
  \begin{tabular}{@{}llllll@{}}
    \toprule
    \multirow{2}{*}{\textbf{Language}} & \multicolumn{2}{c}{\textbf{Informative}} & \multicolumn{2}{c}{\textbf{Non-Informative}} & \multirow{2}{*}{\textbf{F1}} \\
    \cmidrule(lr){2-3} \cmidrule(lr){4-5}
    & \textbf{Prec} & \textbf{Rec} & \textbf{Prec} & \textbf{Rec} & \\
    \cmidrule(r){1-1} \cmidrule(l){6-6} 
    English \((n=2,388)\) & 0.68 & \textbf{0.73} & \textbf{0.86} & 0.83 & \textbf{0.78} \\
    Swahili \((n=87)\) & 0.68 & 0.52 & 0.78 & \textbf{0.88} & 0.71 \\
    \bottomrule
  \end{tabular}
  \caption{Performance Across Language Groups}
  \label{tab:langauge_fairness_1}
\end{table}

\section{Error Analysis}
To better understand the sources of errors in the classification tasks, we conduct an error analysis using a similar approach to \citet{vidgen_detecting_2020}. We systematically explore themes as they emerge from a sample of the misclassified election reports, organise them into categories, and iterate over this process until a `saturated' point where all the data fits into a set of mutually exclusive and exhaustive categories \cite{corbin_grounded_1990}. The errors are broadly categorised into \textbf{annotator errors} and \textbf{machine learning errors}.

\paragraph{Annotator Errors (26\%)} In a few of the cases, the election reports were clearly assigned labels that were inconsistent with the annotation guidelines. Note that this does not mean that 26\% of the dataset is wrongly labelled as this sample is biased by the fact that it has been selected specifically because the model made an `error'. 
\paragraph{Machine Learning Errors (74\%)} The largest proportion of the misclassified informative reports were due to model errors. We further divide them into clear errors (i.e., errors easily identifiable by humans) and edge-case errors (where election reports contain nuances such that the model's misclassification has some merit). 
\begin{description}
\item[Clear Error (Lexical Similarity), 31\%] In several cases, misclassified election reports were clearly assigned the wrong labels. This suggests possible overfitting as the reports that were often mislabelled were lexically similar to the reports which belong to that category, especially if they contained certain keywords. For example, the use of words such as ``hurt'', ``mysteriously'' and ``ghost'' resulted in reports misclassified as \textit{Security Issues} or keywords such as ``win'' or ``stolen'' resulted in reports being misclassified as \textit{Counting and Results}.
\item[Edge Case (Ambiguous Categories), 20\%] of the misclassified  election reports were \textit{Positive Events}. \textit{Positive Event} information category is applied to a wide range of reports that comment on positive aspects of various electoral processes i.e. 
 voting, counting and security and reports containing positive messages like calls for peaceful elections or patriotic rallying. As a result of this, \textit{Positive Events} often contain reports that are conceptually similar to all the other categories making it difficult for the models to distinguish them.
\item[Edge Case (Co-present primary categories), 13\%] A few of the misclassified reports contain multiple information categories whereas the classification tasks are modelled as a single-label task. In such cases, the models often identified one of the co-present categories.
\item[Edge Case (Political Context Nuances), 5\%] of election reports that contained keywords or mentioned locations that required knowledge of the local political context to classify were often misclassified. Unfortunately, such reports are rare in the dataset, leading the model to overlook associations linked to local political nuances.
\item[Edge Case (Multimodal Context), 5\%] The model exhibited frequent misclassifications of election reports that included multimodal content, which could have otherwise provided additional context to the text's meaning. Particularly, reports labelled as \textit{Political Rallies} were commonly misclassified, as many of them had accompanying images or videos.
\end{description}

\subsection{Addressing Classification Errors}
To address model errors stemming from lexical similarities across information categories, likely caused by overfitting on a limited dataset, we recommend collecting a larger, balanced dataset and retraining the models to improve generalization. Furthermore, our analysis is confined to text; exploring multimodal models that incorporate both textual and visual elements of election reports may potentially enhance model performance.  To deal with ambiguous information categories such as \textit{Positive Events}, we suggest an annotation strategy that categorises reports based on their informational content, followed by the application of sentiment analysis to ascertain the positive or negative nature of these reports.

\section{Discussion}
Crowdsourced election reports often present challenges due to noisy, multilingual and code-switched short text sequences that lead to ambiguity and data sparsity in text classification tasks. To mitigate this, we propose a two-step classification approach that first identifies informative election reports and then classifies them into distinct information types. We examine the performance of state-of-the-art multilingual transformer models such as XLM-R and multilingual sentence embeddings on the two classification tasks. The XLM-R model achieves the best performance on the informativeness detection task with a 77\% F1 score, while multilingual Sentence-BERT (SBERT) embeddings on the information type classification task, achieving a 75\% F1 score. Incorporating linguistically motivated features such as the context of an election report, temporal information and sentiment features yielded performance improvements.  Given limited training data and the imperative to create accurate classifiers within a constrained development budget, our findings suggest that multilingual SBERT embeddings provide a viable alternative to fine-tuning a multilingual transformer model without sacrificing performance significantly.

Crowdsourced election monitoring technologies are increasingly deployed to new electoral domains where access to labelled data at the onset of an election monitoring cycle is infeasible or very limited. Scaling crowdsourced election monitoring necessitates developing classifiers that can be transferred to new electoral domains. We stimulate two deployment scenarios: zero-shot, where classifiers are applied to a new electoral domain without any labelled instances, and few-shot, where classifiers are deployed to a target electoral domain with a few labelled reports from the target domain for training. In the zero-shot setting, the XLM-R model fine-tuned on the Kenyan election dataset showed promising results in transferring learning acquired in one electoral domain to another. In the few-shot setting, fine-tuning an XLM-R model, originally trained in a source electoral domain with a few reports from the target electoral domain led to substantial model improvements compared to fine-tuning a domain-specific model anew with equivalent data size. 

Crowdsourced election monitoring has been viewed as a way to mitigate coverage and diversity biases in traditional election monitoring. However, applying such models without considering performance disparities across language groups can further silence certain voices from conversations on electoral integrity. The best-performing model performs better in identifying informative English reports compared to Swahili, largely attributed to language group imbalances in the training data. Deploying such linguistically unfair models without addressing language group imbalances in the training data may result in unintended biases that reinforce and reproduce social hierarchies while failing to recognise marginalised and underrepresented communities \cite{blodgett_language_2020, lauscher_welcome_2022, dev_harms_2021}.  

\section{Conclusion and Future Work}
In this paper, we advance the task of classifying crowdsourced election reports to multilingual and cross-domain classification settings. We adopt a two-step classification approach of first identifying informative election reports and then classifying them into distinct information types. Utilising multilingual transformer-based models enhanced with linguistically motivated features, we achieve notable results in both classification tasks. Our study shows promising results in applying classifiers trained on one electoral domain to a different electoral domain with no or minimal training instances. While our models achieve decent performance, recent developments in Large Language Models (LLMs) suggest the potential for further improvements, particularly in zero-shot text classification tasks. In future work, we aim to explore the efficacy of LLMs in classifying crowdsourced election reports. Given the surge in misinformation observed in recent elections, verifying crowdsourced reports before publication remains a key responsibility of digital response volunteers. However, this task becomes more manageable now, as they only need to verify informative election reports.

\section*{Acknowledgements}
We wish to acknowledge the significant contributions of the Ushahidi team, including Angela Oduor, Daniel Odongo, and Joseph Kirai, for granting access to the crowdsourced election datasets and providing valuable feedback during this project. We acknowledge the Rhodes Scholarship for supporting Jabez Magomere's studies at the University of Oxford. 

\bibliography{aaai24}

\appendix
\section{Exploratory Data Analysis}
\begin{figure}[htbp]
    \centering
    \begin{minipage}{\columnwidth}
        \centering
        \centering
        \includegraphics[width=\columnwidth]{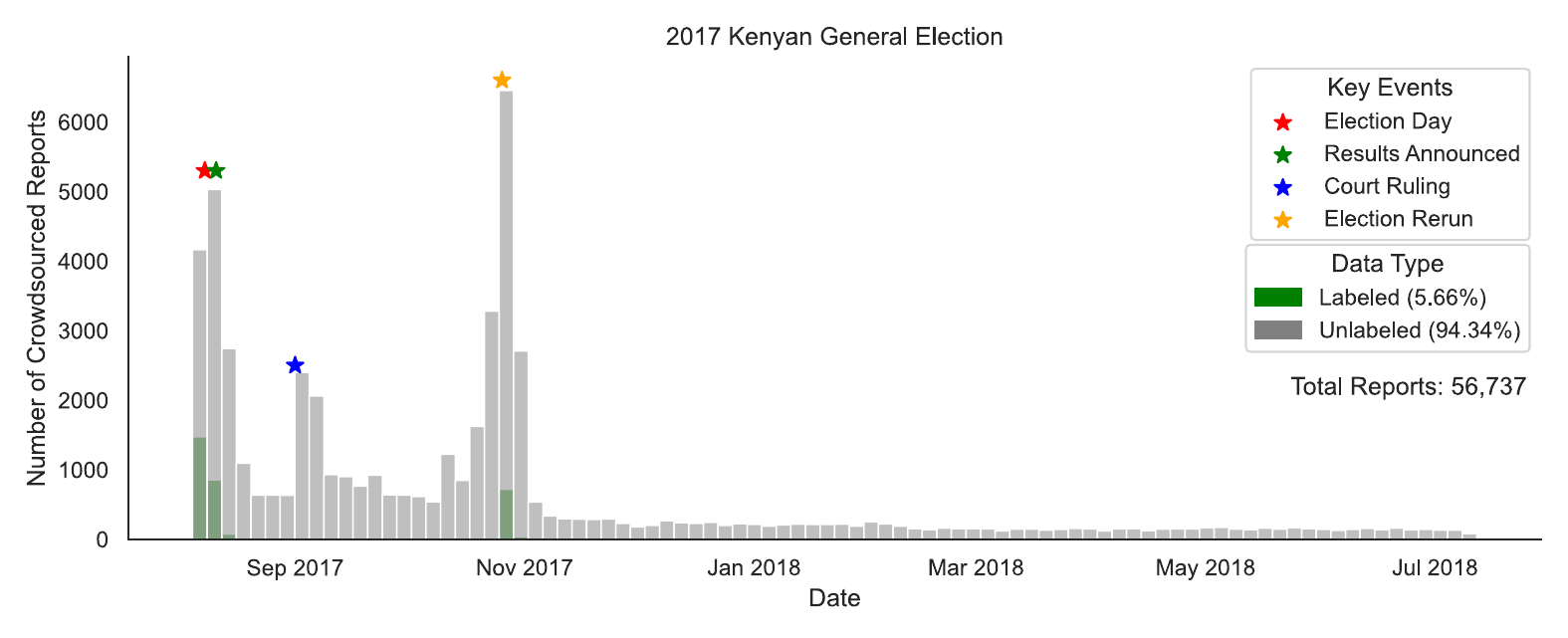} 
    \end{minipage}
    \begin{minipage}{\columnwidth}
        \centering
        \includegraphics[width=\columnwidth]{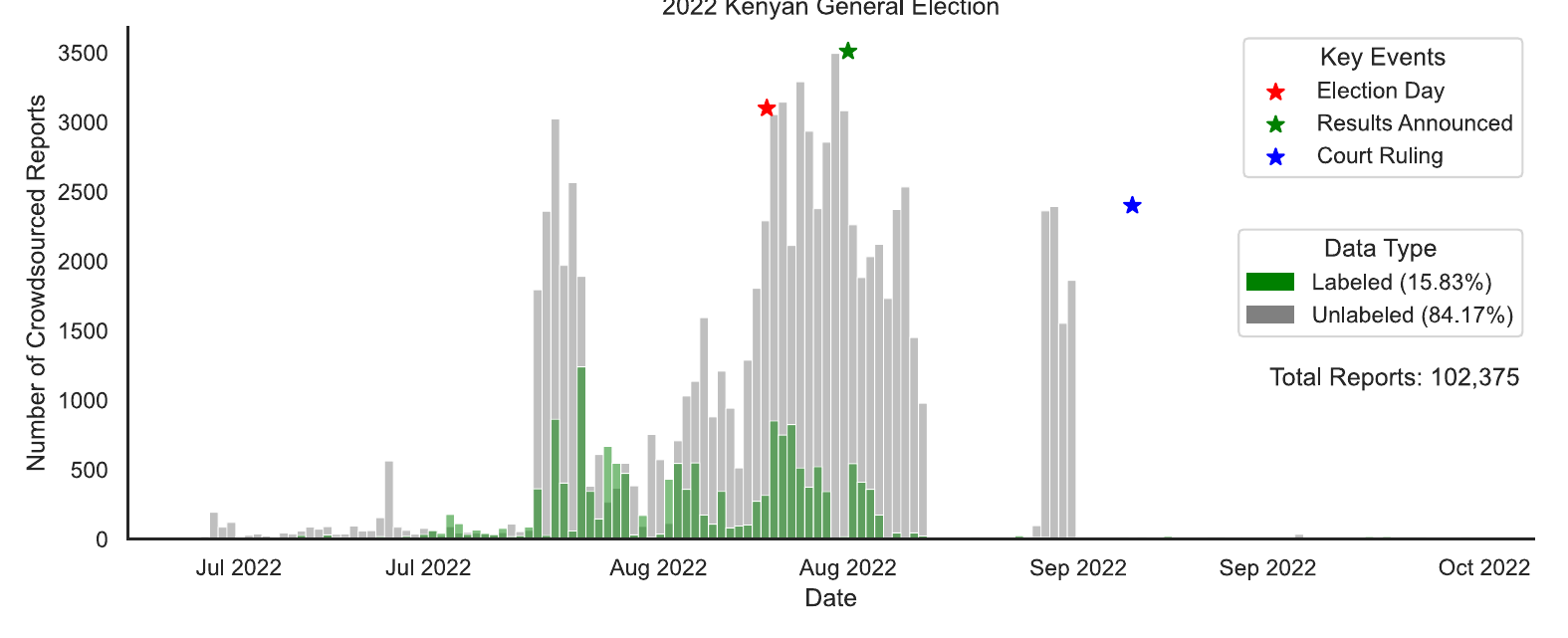}
    \end{minipage}

    \begin{minipage}{\columnwidth}
        \centering
        \includegraphics[width=\columnwidth]{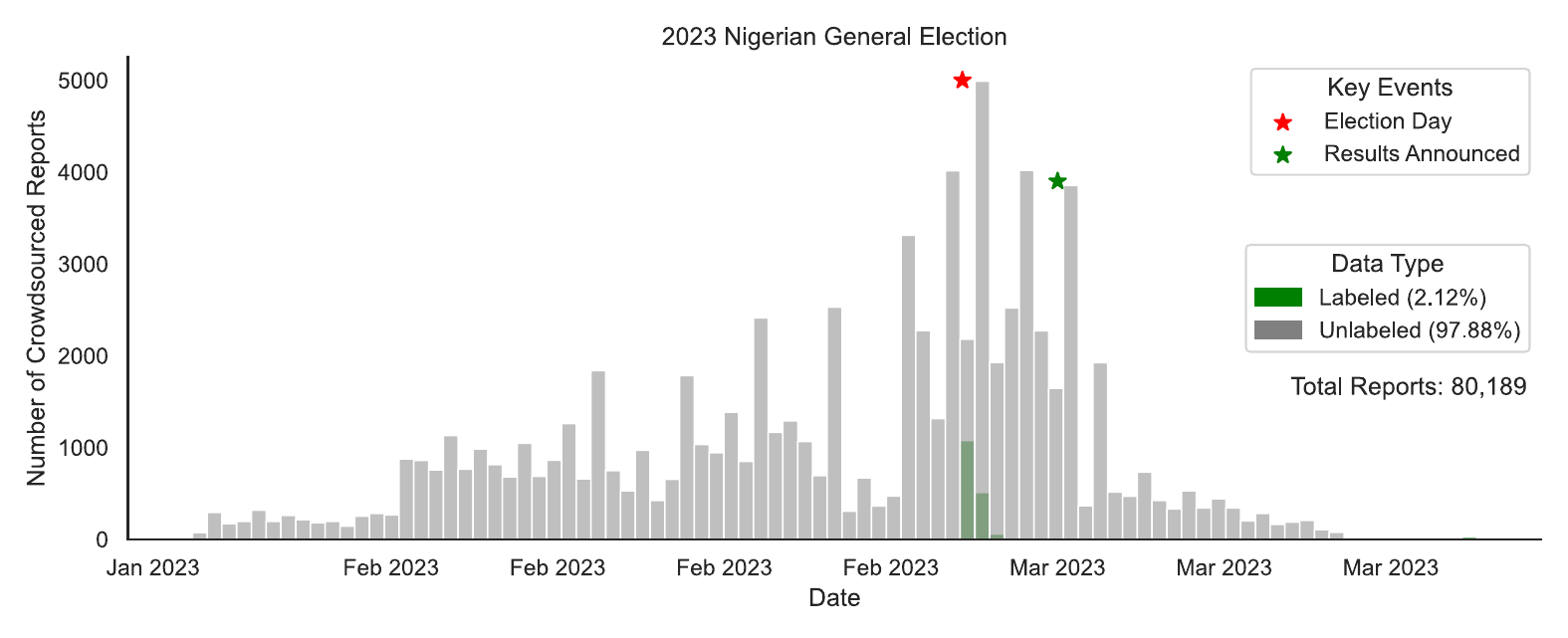}
    \end{minipage}
    \caption{Timelines of Crowdsourced Election Monitoring Deployments: 2017 and 2022 Kenyan General Elections, and 2023 Nigerian General Election.}
    \label{fig:time_series}
\end{figure}

Figure \ref{fig:time_series} highlights deployment timelines for crowdsourced election monitoring technologies during the 2017 and 2022 Kenyan General Elections and the 2023 Nigerian General Election. The first deployment in the dataset ran between July 2017 and July 2018, during the 2017 Kenyan General Election and crowdsourced  56,737 election reports. The second deployment took place between July 2022 and October 2022 encompassing the 2022 Kenyan General Election and crowdsourced 102,375 reports. The third deployment spanning between December 2022 and March 2023 encompasses the 2023 Nigerian General Election crowdsourcing a total of 80,207 election reports. 

Notably, peaks in the number of crowdsourced reports align with critical events in the electoral cycle consistent across all election deployments. For instance, the highest number of reports are generated between the election day and the announcement of results, with additional peaks corresponding to other key electoral events such as court rulings on election outcomes, as observed in the 2017 Kenyan General Election \cite{freytas-tamura_kenya_2017}. 

The distribution of labelled reports primarily centres around the election month, as shown in Figure \ref{fig:time_series}. Across all election deployments, unlabelled crowdsourced election reports significantly outnumber labelled reports. For the 2017 Kenyan General Election, approximately 6\% (3,209 out of 56,737 reports) of the reports are labelled. Similarly, during the 2022 Kenyan General Election, around 16\% (16,208 out of 102,375 reports) of the reports were labelled. The trend persists in the Nigerian 2023 Election, where only 2\% (1,700 out of 78,507) of the election reports are labelled.
\section{List of Information Categories}
\label{subsec:info_categories}
Table \ref{tab:information_categories} shows the list of information categories and descriptions applied by \textit{Uchaguzi} in the Kenyan Election deployments. Digital response volunteers are required to assign an information category to each incoming election report.
\begin{table}[!hbp]
\centering
\begin{tabularx}{\columnwidth}{@{}XX@{}}
\toprule
\textbf{Information Category} & \textbf{Description} \\
\midrule
Opinions or Others & Political Sentiments, Subjective Opinions, Jokes, Trolls \\
Political Rallies & Political Campaigns, Political Messaging, Election Promises \\
Positive Events & Civilian Peace Efforts, Everything Fine, Police Peace Efforts \\
Security Issues & Mobilisation towards Violence, Threat of Violence, Dangerous Speech, Armed Clashes, Riots \\
Voting Issues & Voter Registration irregularities, Voter Integrity Irregularities, Purchasing of Voters Cards \\
Staffing Issues & Absence or insufficient number of election officials, Election Officials not acting according to rules \\
Polling Station Administration & Polling station logistical issues, Missing/Inadequate Voting Materials, Ballot Box Irregularities \\
Counting and Results & Failure to announce results, Irregularities with transportation of ballot boxes, Counting Irregularities, Party Agent Irregularities, Provisional Citizen Result \\
Media Reports & Observations from established media houses \\
\bottomrule
\end{tabularx}
\caption{Information Categories and Descriptions applied by Uchaguzi \protect\cite{ushahidi_survey_2020}.}
\label{tab:information_categories}
\end{table}

\section{Distribution of Crowdsourced Reports Distance from Election Day}
\label{sec:dist_appendix}
\begin{figure}[!h]
    \centering
    \includegraphics[width=\columnwidth]{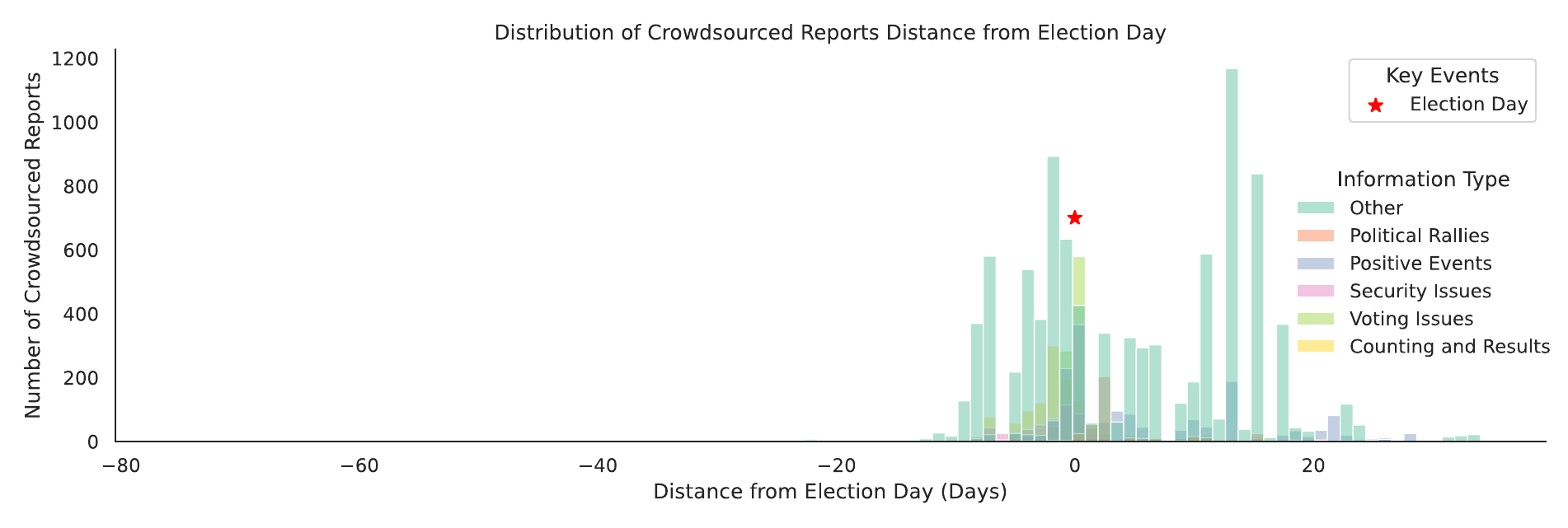}
    \caption{Distribution of Crowdsourced Reports Distance from Election Day.}
    \label{fig:election_distance}
\end{figure}

Crowdsourced election reports exhibit a temporal dependency, with election reports of different information types being received at different times, as shown in Figure \ref{fig:election_distance}.

\section{Text Pre-Processing}
Crowdsourced election reports are inherently noisy, and cleaning the text before further processing helps to generate better features. The selection of these techniques is based on model-specific considerations and initial experimental results. The following pre-processing steps are applied in the study:
\begin{itemize}
    \item All Unicode characters, extra white spaces, and punctuation are removed from the text.
    \item All occurrences of URLs, user mentions, and numbers are removed.
    \item Repeated full stops, question marks, and exclamation marks are removed.
    \item All election-specific hashtags, such as \#KenyaDecides2022, \#electionske, and \#KenyaDecides, are excluded. However, the study observed that some additional information was included in the election report in the form of hashtags, and maintaining them may be helpful for the classification tasks. Therefore, all remaining hashtags were retained with only the hashtag symbol (\#) removed in front of the word.
    \item All English stopwords are removed using NLTK's\footnote{Natural Language Toolkit Python Package} stopwords list.
    \item Tokenization is performed using NLTK's TweetTokenizer due to the prevalence of tweets in the dataset.
    \item Finally, all tokens are converted to lowercase.
\end{itemize}
After initial experiments, it was observed that the multilingual transformer-based models performed better with minimal text preprocessing. 
The only pre-processing steps applied for transformer-based models were replacing all user mentions with a \texttt{USR} token, URLs with the \texttt{HTTP} token, and converting emojis to their corresponding text representations using the Python library emoji\footnote{\url{https://pypi.org/project/emoji/}}.

\section{Model Implementation and Hyperparameter Tuning}
\label{sec:tfidf_appendix}
\begin{figure}[!h]
    \centering
    \includegraphics[width=\columnwidth]{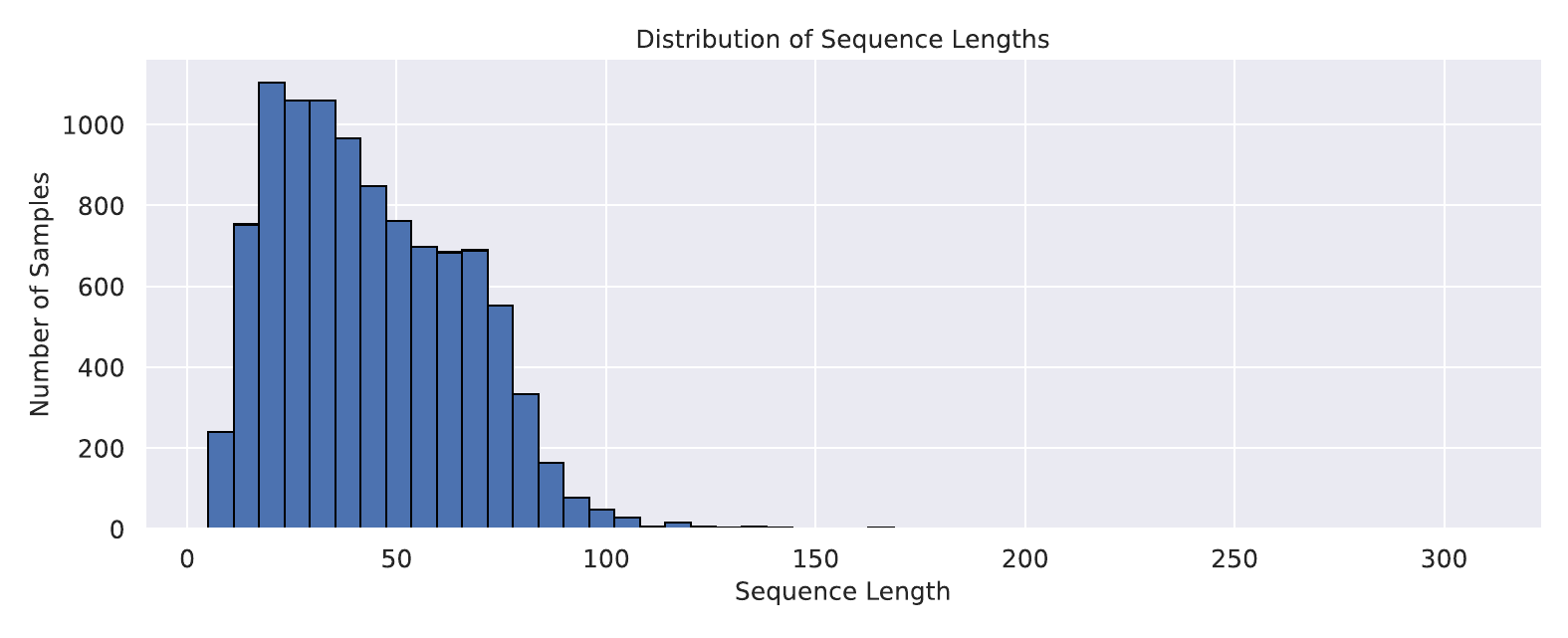}
    \caption{Distribution of crowdsourced election reports sequence Lengths.}
    \label{fig:seq_distribution}
\end{figure}

\begin{figure}[!h]
    \centering
    \includegraphics[width=\columnwidth] {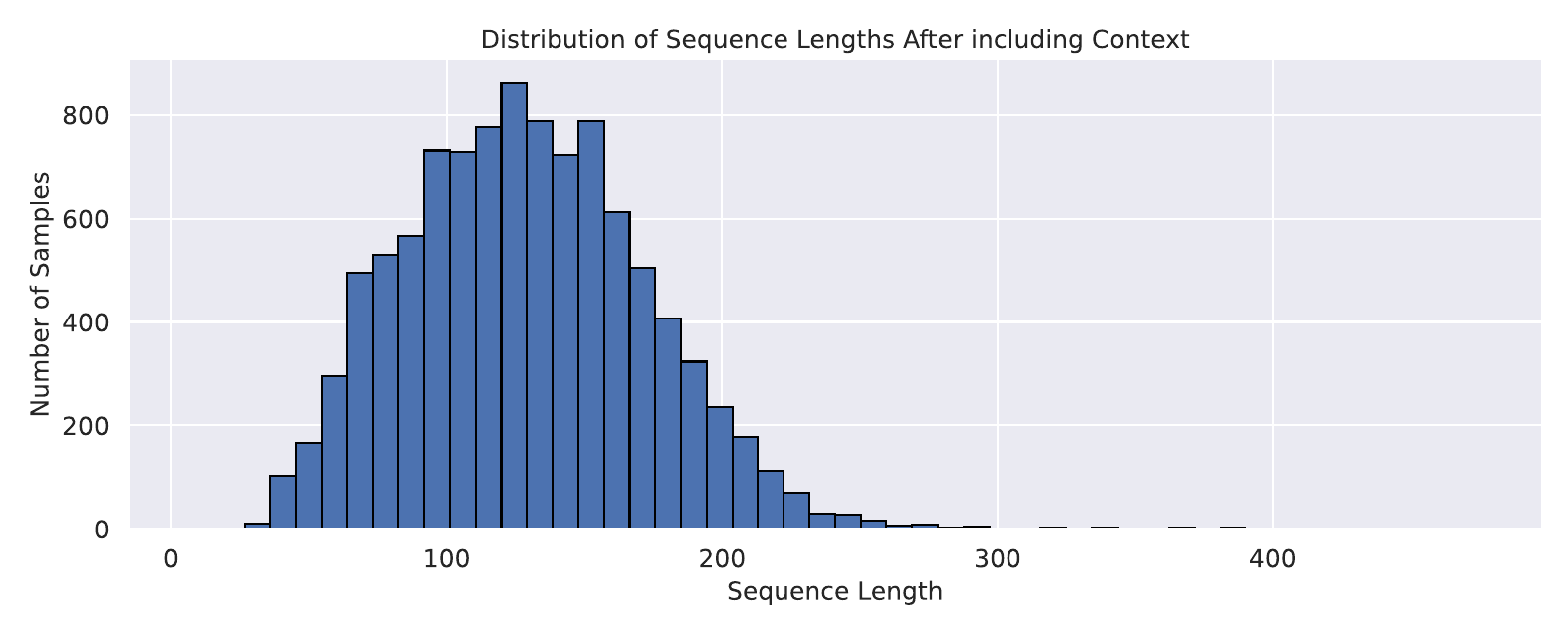}
    \caption{Distribution of Sequence Lengths after including context.}
    \label{fig:seq_distribution_with_context}
\end{figure}

We use the Sentence Transformers Library \footnote{\url{https://huggingface.co/sentence-transformers}} provided by Hugging Face to map election reports text into 768-dimensional dense vectors. We fine-tune the transformer-based models for the text classification tasks using the Hugging Face Transformers library\footnote{\url{https://huggingface.co/docs/transformers}} with the help of PyTorch\footnote{\url{https://pytorch.org/}}.  The cleaned election reports are tokenised using off-the-shelf BERT-based tokenisers for each model. We set a maximum sequence length of \(128\) before including the context and \(300\) when we include the context, based on an analysis of the sequence length distributions shown in Figure \ref{fig:seq_distribution} and Figure \ref{fig:seq_distribution_with_context}. 
We conducted training using the same hyperparameter sweep identified in \citet{liu_robustly_2021} as the most effective for the GLUE benchmark tasks. This includes testing across learning rates $\in \{1e-5, 2e-5, 3e-5\}$ and batch sizes $\in \{16, 32\}$. Each model undergoes training for \(5\) epochs, although after conducting initial experiments on the development set, the models trained for binary classification tasks are trained for \(2\) epochs, while those for the multiclass classification task are trained for \(4\) epochs. To optimise model performance, we utilise the AdamW \cite{loshchilov_decoupled_2017} algorithm and implement a scheduler that incorporates linear warmup and decay.

To deal with class imbalances, we apply class weights to the cross-entropy loss function. These weights are estimated using Scikit-Learn's compute class weight function, whereby minority classes are assigned higher weights. During training, the model incurs higher penalties for misclassifying minority labels, compelling it to learn the distribution of the rarer classes.

\section{Baseline Model Comparison}
\label{sec:tfidf_appendix}
\subsection{Informativeness Detection}
Table \ref{tab:baseline_models_comparison} shows the results of the baseline models on the informativeness detection task. 
\begin{table}[hbt!]
  \small
  \centering
  \begin{tabular}{@{}llllll@{}}
    \toprule
    \thead{Model}                   & \thead{Accuracy} & \thead{Precision} & \thead{Recall} & \thead{F1} \\
    \midrule
    SVM                &  \textbf{0.79}          & 0.74           & 
    \textbf{0.79} &   \textbf{0.75 }          \\
    Logistic Regression           & 0.76        & 0.74         & 0.74     & 0.74          \\
    Naive Bayes                  & 0.77            & 0.74        & 0.75       & 0.75        \\
    \bottomrule
  \end{tabular}
  \caption{Comparison Baseline Models on the Informativeness Detection task (Maximum columnar values in bold)}
  \label{tab:baseline_models_comparison}
\end{table}

\subsection{Information Types Classification}
Table \ref{tab:baseline_models_comparison_it} shows the results of the baseline models on the information types classification task. 
\begin{table}[hbt!]
  \small
  \centering
  \begin{tabular}{@{}llllll@{}}
    \toprule
    \thead{Model}                   & \thead{Accuracy} & \thead{Precision} & \thead{Recall} & \thead{F1} \\
    \midrule
    SVM                &  \textbf{0.70}          & \textbf{0.70 }          & 
    0.69 &   \textbf{0.69 }          \\
    Logistic Regression           & 0.69        & 0.68         & 0.70     & 0.69          \\
    Naive Bayes                  & 0.67           & 0.65        & 0.70       & 0.66       \\
    \bottomrule
  \end{tabular}
  \caption{Comparison of Baseline Models on the Information Types Classification Task (Maximum columnar values in bold)}
  \label{tab:baseline_models_comparison_it}
\end{table}

\section{BERT-based Models Comparison}
\label{sec:bert_appendix}
\subsection{Informativeness Detection}
Table \ref{tab:bert_models_comparison} shows the comparison of XLM-R \cite{conneau_unsupervised_2019} and mDeBERTav3 \cite{he_debertav3_2021} models on the informativeness detection task. 
\begin{table}[hbt!]
  \small
  \centering
  \begin{tabular}{@{}llllll@{}}
    \toprule
    \thead{Model}                   & \thead{Accuracy} & \thead{Precision} & \thead{Recall} & \thead{F1-Score} \\
    \midrule
    XLM-R                &  \textbf{0.79}          & \textbf{0.77}         & 
    \textbf{0.78} &   \textbf{0.77 }          \\
    mDeBERTa-v3           & 0.78        & 0.76         & 0.77     & 0.76          \\
    \bottomrule
  \end{tabular}
  \caption{Comparison of BERT models performance on the Informativeness Detection task (Maximum columnar values in bold)}
  \label{tab:bert_models_comparison}
\end{table}

\subsection{Information Types Classification}
Table \ref{tab:bert_models_comparison_it} shows the comparison of XLM-R, mDeBERTav3 and mBERT \cite{devlin_bert_2018} models on the information types classification task. 
\begin{table}[hbt!]
  \small
  \centering
  \begin{tabular}{@{}llllll@{}}
    \toprule
    \thead{Model}                   & \thead{Accuracy} & \thead{Precision} & \thead{Recall} & \thead{F1-Score} \\
    \midrule
    XLM-R                &  \textbf{0.70}          & 0.68         & 
    \textbf{0.73} &   \textbf{0.70}          \\
    mDeBERTa-v3           & 0.69        & 0.68         & 0.70     & 0.69          \\
    mBERT          & 0.66        & 0.67         & 0.70     & 0.68          \\
    \bottomrule
  \end{tabular}
  \caption{Comparison of BERT models performance on the Information Types Classification task (Maximum columnar values in bold)}
  \label{tab:bert_models_comparison_it}
\end{table}

\end{document}